\documentclass{article}
\usepackage{spconf,amsmath,amssymb,graphicx,booktabs,hyperref}
\hypersetup{hidelinks}

\title{REH-FUSE: RELIABILITY-AWARE HIERARCHICAL FUSION OF EXPERTS FOR MULTIMODAL EMOTION RECOGNITION IN CONVERSATION}

\name{Guan-Hua Wen  \qquad Hou-Chiang Tseng  \qquad Kuan-Yu Chen }
\address{National Taiwan University of Science and Technology, Taipei, Taiwan}

\begin{document}
\maketitle

\begin{abstract}
Multimodal emotion recognition in conversation (ERC) requires adapting to the instance-dependent reliability of different evidence sources. Lexical content may be decisive, vocal expression may provide complementary cues, or accurate recognition may require cross-modal interaction; fixed fusion does not explicitly account for this variation. We propose ReH-FUSE, a reliability-aware framework with dialogue-aware text, audio, and cross-modal experts. Its decision-level router first models the relative preference between text and audio and then balances the resulting unimodal mixture against the cross-modal expert. This factorization separates unimodal competition from cross-modal selection. Across three independent runs on IEMOCAP, ReH-FUSE achieves $74.34\%$ weighted F1 and $73.11\%$ macro F1; on MELD, it achieves $68.03\%$ weighted F1. Controlled ablations show that learned routing outperforms uniform expert averaging and benefits from cross-modal interaction.
\end{abstract}


\section{Introduction}
\label{sec:introduction}

Emotion recognition in conversation (ERC) predicts an emotion for each utterance while accounting for dialogue history, speaker interaction, commonsense, and emotion transitions \cite{dialoguernn,dialoguegcn,cosmic,dagerc}. Multimodal ERC additionally uses non-lexical evidence, most commonly speech. Its central difficulty is that evidence reliability varies by utterance: text may directly reveal an emotion, vocal expression may disambiguate neutral wording, or neither modality may suffice without their interaction. Consequently, the problem is not merely how to combine modalities, but which evidence path should drive each prediction.

Many multimodal systems combine representations at predetermined stages or treat cross-modal interaction as uniformly beneficial \cite{mmgcn,arevalo2017gmu}. This assumption can propagate noisy evidence, dilute an already sufficient unimodal cue, and fail to express which source should contribute most strongly when text and audio disagree. Strong pretrained encoders improve candidate representations but do not solve evidence selection. We therefore formulate multimodal ERC as routing over competing decision sources.

We propose \textbf{ReH-FUSE} (Reliability-Aware Hierarchical Fusion of Experts), comprising contextualized text, audio, and cross-modal predictors. Instead of a flat three-way gate, its router first models the relative preference between the unimodal experts and then balances their mixture against cross-modal interaction. This decomposition mirrors two distinct questions: which unimodal source is more useful, and whether unimodal evidence is sufficient.

Our contributions are threefold. First, we formulate multimodal ERC as instance-dependent routing over evidence sources with varying reliability. Second, we introduce a hierarchical decision-level router that separates unimodal competition from cross-modal selection. Third, repeated evaluations, matched architectural ablations, and gate analyses demonstrate the benefit of learned expert routing over uniform fusion.

\section{Related Work}
\label{sec:related}

Context-aware ERC models use recurrent state tracking, graph message passing, commonsense, and directed dialogue structures to model speakers and emotion transitions \cite{dialoguernn,dialoguegcn,cosmic,dagerc}. Multimodal benchmarks such as IEMOCAP and MELD established the complementary roles of semantic and expressive cues \cite{iemocap,meld}; subsequent methods introduced graph interaction, hierarchical cross-attention, disentangled context-modality representations, and teacher-led fusion \cite{mmgcn,dutta-ganapathy-2023-hcam,li-etal-2023-df-erc,yun-etal-2024-telme}.

Adaptive expert models learn input-dependent computation \cite{jacobs1991adaptive,shazeer2017outrageously,fedus2022switch}. Gated multimodal units similarly modulate modality contributions \cite{arevalo2017gmu}, while MiSTER-E applies decision-level MoE to ERC \cite{dutta-etal-2026-mistere}. ReH-FUSE differs in factorizing the routing decision: modality-specific dialogue modeling precedes cross-modal interaction, and text-versus-audio preference is separated from unimodal-versus-cross-modal selection. The gate can therefore be interpreted as structured competition among contextualized decision sources rather than adaptive feature interpolation.

\section{ReH-FUSE}
\label{sec:method}

\begin{figure}[t]
 \centering
 \includegraphics[width=\linewidth]{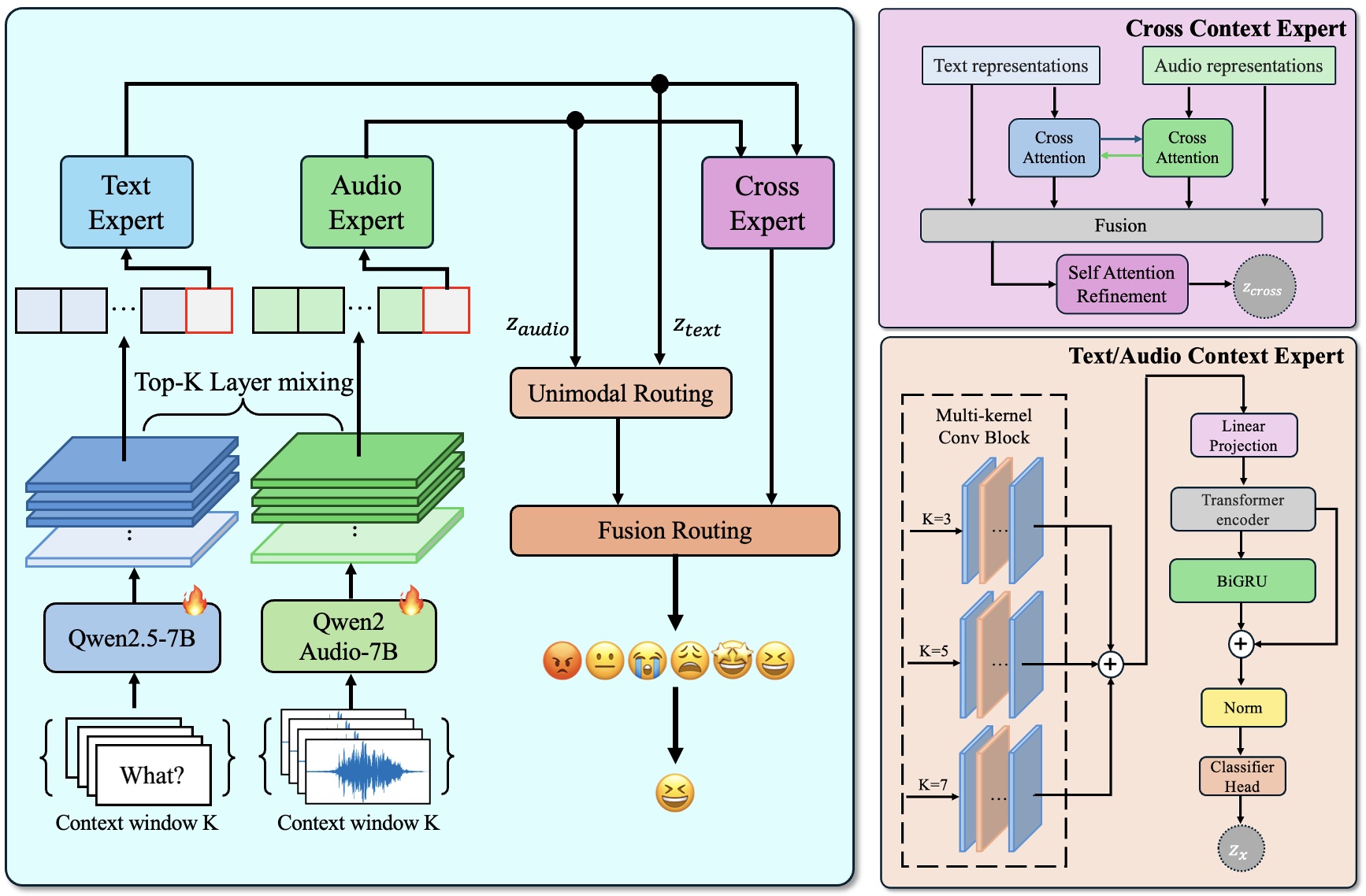}
 \caption{ReH-FUSE architecture with text, audio, and cross-modal experts followed by hierarchical decision routing.}
 \label{fig:architecture}
\end{figure}

\subsection{Problem formulation}

Figure~\ref{fig:architecture} summarizes the three expert paths and hierarchical decision router. Let a dialogue be $\mathcal{D}=\{(u_i,a_i,s_i)\}_{i=1}^{N}$, where $u_i$, $a_i$, and $s_i$ are the text, aligned audio, and speaker of utterance $i$. Given a context window
\begin{equation}
 \mathcal{H}_i=\{(u_j,a_j,s_j)\}_{j=\max(1,i-K)}^{i},
\end{equation}
the task is to predict $y_i\in\{1,\ldots,C\}$. Three experts produce logits $\mathbf{z}_i^{(t)}$, $\mathbf{z}_i^{(a)}$, and $\mathbf{z}_i^{(c)}$ for text, audio, and cross-modal evidence. The routed prediction is
\begin{equation}
 \mathbf{z}_i=\sum_{e\in\{t,a,c\}}g_i^{(e)}\mathbf{z}_i^{(e)},\quad
 \sum_e g_i^{(e)}=1.
 \label{eq:routed-prediction}
\end{equation}
The weights $\mathbf{g}_i$ specify source contributions. Here, reliability denotes learned routing preference, not calibrated signal quality.

\subsection{Utterance encoding and context experts}

The text input concatenates the current utterance with at most four preceding turns using compact turn and local-speaker markers. Qwen2.5-7B encodes this sequence with LoRA adaptation \cite{yang2024qwen2,lora}. Audio is converted to mono at 16~kHz, truncated to 6~s on IEMOCAP or 8~s on MELD, and encoded with Qwen2-Audio-7B \cite{chu2024qwen2audio}. For both modalities, a learned mixture of the top four hidden layers is pooled at the last valid step and projected to a shared hidden dimension. Learned speaker embeddings are concatenated after projection.

Before modalities interact, separate text and audio context experts process each dialogue sequence. Each expert combines multi-kernel temporal convolutions, a Transformer encoder, and a bidirectional GRU with residual normalization:
\begin{equation}
 \mathbf{H}_t=f_{\mathrm{ctx}}^{(t)}(\mathbf{X}_t),\qquad
 \mathbf{H}_a=f_{\mathrm{ctx}}^{(a)}(\mathbf{X}_a).
 \label{eq:context}
\end{equation}
This arrangement captures local patterns, longer dependencies, and conversational flow while preserving modality-specific evidence.

\subsection{Cross-modal expert}

The cross-modal branch applies attention in both directions. Text queries audio and audio queries text, after which an MLP fuses the original and attended sequences:
\begin{align}
 \widetilde{\mathbf{H}}_t&=\mathrm{Attn}(\mathbf{H}_t,\mathbf{H}_a), &
 \widetilde{\mathbf{H}}_a&=\mathrm{Attn}(\mathbf{H}_a,\mathbf{H}_t),\\
 \mathbf{H}_c&=f_{\mathrm{fuse}}([\mathbf{H}_t,\mathbf{H}_a,
 \widetilde{\mathbf{H}}_t,\widetilde{\mathbf{H}}_a]).
\end{align}
A self-attention refinement block then yields the cross-modal representation. Separate classifiers map $\mathbf{H}_t$, $\mathbf{H}_a$, and $\mathbf{H}_c$ to the three expert logits in Eq.~\eqref{eq:routed-prediction}.

\subsection{Hierarchical decision routing}

The router consumes contextual representations and pre-softmax expert predictions. Its first-stage input is $\mathbf{q}_i^{(m)}=[\mathbf{H}_{t,i},\mathbf{H}_{a,i},\mathbf{z}_i^{(t)},\mathbf{z}_i^{(a)}]$. An MLP and softmax produce $\mathbf{m}_i=[m_i^{(t)},m_i^{(a)}]$, with $m_i^{(t)}+m_i^{(a)}=1$. We form $\mathbf{H}_{u,i}=m_i^{(t)}\mathbf{H}_{t,i}+m_i^{(a)}\mathbf{H}_{a,i}$ and $\mathbf{z}_i^{(u)}=m_i^{(t)}\mathbf{z}_i^{(t)}+m_i^{(a)}\mathbf{z}_i^{(a)}$. The second-stage input is $\mathbf{q}_i^{(f)}=[\mathbf{H}_{u,i},\mathbf{H}_{c,i},\mathbf{z}_i^{(u)},\mathbf{z}_i^{(c)}]$. A second MLP and softmax produce $\mathbf{f}_i=[f_i^{(u)},f_i^{(c)}]$, which allocates weight between the unimodal mixture and cross-modal expert. Their structured three-way distribution is
\begin{equation}
 \mathbf{g}_{i,\mathrm{base}}=
 [f_i^{(u)}m_i^{(t)},\ f_i^{(u)}m_i^{(a)},\ f_i^{(c)}].
 \label{eq:hierarchical-gate}
\end{equation}
Final weights are obtained by applying softmax to scaled log probabilities, with an optional global text-prior bias. The evaluated recipe does not append uncertainty statistics or engineered reliability features to $\mathbf{q}_i^{(m)}$ or $\mathbf{q}_i^{(f)}$. In contrast to flat routing, Eq.~\eqref{eq:hierarchical-gate} explicitly encodes the two decisions underlying evidence selection. Training uses focal loss \cite{lin2017focal} on the routed logits and an entropy regularizer that discourages premature gate collapse without forcing uniform routing.

The factorization is deliberately imposed at the decision level. Feature gates may suppress a modality before it forms a contextualized prediction, whereas a flat decision gate must learn unimodal competition and cross-modal necessity jointly. ReH-FUSE preserves three complete hypotheses: $\mathbf{m}_i$ describes modality preference and $f_i^{(c)}$ describes demand for interaction. Soft weights keep every expert differentiable and express intermediate uncertainty rather than irreversible assignments.

\section{Experiments}
\label{sec:experiments}

\subsection{Datasets and protocol}

IEMOCAP contains dyadic interactions with six labels; MELD contains multi-party conversations with seven labels \cite{iemocap,meld}. We use their standard splits and select checkpoints by validation W-F1. Qwen2.5-7B and Qwen2-Audio-7B serve as the text and audio backbones, with a learned top-four layer mixture, last-step pooling, 256 text tokens, $K=4$, batch size 1, and gradient accumulation 4. Audio is truncated to 6~s for IEMOCAP and 8~s for MELD. Main results and matched IEMOCAP ablations are reported over three random seeds.

Table~\ref{tab:configuration} summarizes shared settings; only maximum audio duration differs between datasets. Text prompts expose no future utterances.

\begin{table}[t]
 \centering
 \caption{Shared training and input configuration.}
 \label{tab:configuration}
 \small
 \setlength{\tabcolsep}{3.5pt}
 \begin{tabular}{lc}
  \toprule
  Component & Setting \\
  \midrule
  Text / audio backbone & Qwen2.5-7B / Qwen2-Audio-7B \\
  Text adaptation & LoRA \\
  Hidden-state aggregation & Learned top-4 mixture \\
  Utterance pooling & Last valid step \\
  Text length / history & 256 tokens / $K=4$ \\
  Audio duration & 6 s (IEMOCAP), 8 s (MELD) \\
  Batch / accumulation & 1 / 4 \\
  Checkpoint criterion & Validation W-F1 \\
  \bottomrule
 \end{tabular}
\end{table}

Router targets are latent because no annotation identifies which expert to prefer. Initial preference may therefore reflect logit scale rather than evidence quality. Warm-up establishes usable experts before joint optimization; temperature controls routing sharpness and entropy regularization helps prevent premature gate collapse. Test metrics are not used for selection.

\subsection{Main results}

Table~\ref{tab:main-results} compares ReH-FUSE with representative context-aware and multimodal fusion systems. ReH-FUSE achieves $74.34\%$ W-F1 on IEMOCAP, an absolute improvement of $3.44$ points over the strongest listed comparator. On MELD, it achieves $68.03\%$ W-F1, remaining competitive with prior systems but below MiSTER-E.

\begin{table}[t]
 \centering
 \caption{Models are compared in weighted F1 (\%) on IEMOCAP and MELD. $\dagger$ denotes the second-best result.}
 \label{tab:main-results}
 \setlength{\tabcolsep}{3.5pt}
 \begin{tabular}{lcc}
  \toprule
  Method & IEMOCAP & MELD \\
  \midrule
  SMIN \cite{lian-etal-2023-smin} & 70.5 & 63.7 \\
  HCAM \cite{dutta-ganapathy-2023-hcam} & 70.5 & 65.8 \\
  DF-ERC \cite{li-etal-2023-df-erc} & 69.5 & 64.5 \\
    Broad-Mamba \cite{shou-etal-2025-broad-mamba} & 70.2 & 65.6 \\
  TelME \cite{yun-etal-2024-telme} & 69.3 & 67.2 \\
    CFN-ESA \cite{li-etal-2024-cfn-esa} & 68.5 & 65.9 \\
  MiSTER-E \cite{dutta-etal-2026-mistere} & $70.9^{\dagger}$ & $\mathbf{69.5}$ \\
  \midrule
  ReH-FUSE & $\mathbf{74.34}$ & $68.03^{\dagger}$ \\
  \bottomrule
 \end{tabular}
\end{table}

The controls answer complementary questions. Uniform averaging shows that three predictions without selection are insufficient. The adaptive flat gate isolates the smaller benefit of hierarchical factorization. Removing the cross-modal expert tests whether unimodal routing alone explains the result and produces the largest loss. Single-path evaluation is not separately trained unimodal modeling because shared modules still receive the complete objective; the comparison therefore concerns inference routes within one matched architecture.

\subsection{Ablation and routing analysis}

Table~\ref{tab:ablation-results} distinguishes expert availability from routing structure on IEMOCAP. Uniform averaging loses 1.28 W-F1, while a flat gate narrows the gap to 0.34, indicating a modest benefit from hierarchical factorization. Removing the cross-modal branch causes the largest drop. Neither single path recovers the routed result, and full-joint training underperforms the warm-up-then-joint schedule.

\begin{table}[t]
 \centering
 \caption{Matched IEMOCAP architectural ablations (\%).}
 \label{tab:ablation-results}
 \setlength{\tabcolsep}{5pt}
 \begin{tabular}{lcc}
  \toprule
  Configuration & W-F1 & M-F1 \\
  \midrule
  ReH-FUSE & \textbf{74.94} & \textbf{73.53} \\
  w/o hierarchical router & 74.36 & 72.57 \\
  Flat three-way gate & 74.60 & 73.34 \\
  Uniform expert average & 73.66 & 71.70 \\
  w/o cross-modal expert & 66.37 & 63.32 \\
  Text-only path & 74.57 & 73.31 \\
  Audio-only path & 74.42 & 73.20 \\
  Full-joint schedule & 71.34 & 69.41 \\
  \bottomrule
 \end{tabular}
\end{table}

Table~\ref{tab:routing-behavior} provides a behavioral check on the reliability interpretation. Cross-modal routing dominates on average, but weights vary by class. Among unimodal branches, audio outweighs text for angry and happy utterances, whereas text outweighs audio for excited utterances. Neutral, sad, and frustrated receive the largest average cross-modal weights. Thus, routing varies across utterances and classes rather than collapsing to fixed fusion.

\begin{table}[t]
 \centering
 \caption{Mean IEMOCAP routing weights across three runs.}
 \label{tab:routing-behavior}
 \setlength{\tabcolsep}{5.2pt}
 \begin{tabular}{lccc}
  \toprule
  Class & Text & Audio & Cross \\
  \midrule
  Angry & .13 & .21 & .66 \\
  Excited & .26 & .12 & .62 \\
  Frustrated & .12 & .12 & .76 \\
  Happy & .17 & .20 & .63 \\
  Neutral & .08 & .08 & .85 \\
  Sad & .10 & .14 & .76 \\
  \midrule
  Overall & .14 & .13 & .73 \\
  \bottomrule
 \end{tabular}
\end{table}

\subsection{Further analysis}

Table~\ref{tab:stage-experts} shows that all standalone IEMOCAP experts improve after joint training. Cross-modal accuracy gains 30.14 points and remains strongest; text and audio gain 29.53 and 15.29 points. A strong cross-modal expert does not imply unit routing weight: retaining unimodal paths can, in principle, preserve useful decisions when interaction is unnecessary or detrimental.

\begin{table}[t]
 \centering
 \caption{Standalone IEMOCAP expert accuracy (\%).}
 \label{tab:stage-experts}
 \small
 \setlength{\tabcolsep}{6pt}
 \begin{tabular}{lccc}
  \toprule
  Checkpoint & Text & Audio & Cross \\
  \midrule
  Warm-up & 33.29 & 21.76 & 44.27 \\
  Joint training & 62.82 & 37.05 & 74.41 \\
  \bottomrule
 \end{tabular}
\end{table}

\begin{table}[t]
 \centering
 \caption{Per-class F1 (\%) from three-run analyses.}
 \label{tab:classwise-results}
 \small
 \setlength{\tabcolsep}{4pt}
 \begin{tabular}{lcc}
  \toprule
  Class & IEMOCAP & MELD \\
  \midrule
  Angry/anger & 68.7 & 57.6 \\
  Happy/joy & 61.5 & 63.0 \\
  Neutral & 76.7 & 80.5 \\
  Sad/sadness & 82.8 & 44.1 \\
  Frustrated & 71.1 & -- \\
  Excited/surprise & 77.9 & 62.0 \\
  Disgust & -- & 37.2 \\
  Fear & -- & 25.5 \\
  \bottomrule
 \end{tabular}
\end{table}

Aggregate W-F1 can obscure whether gains extend beyond frequent emotions. Table~\ref{tab:classwise-results} shows that IEMOCAP is comparatively balanced: all classes exceed 60 F1, with sad and excited strongest and happy weakest. MELD presents a different failure mode. Neutral and joy remain strong, but rare fear and disgust reach only 25.5 and 37.2, and sadness also trails the aggregate result. The competitive MELD W-F1 should therefore not be interpreted as uniformly reliable recognition across labels.

\begin{figure}[t]
 \centering
 \includegraphics[width=0.82\linewidth]{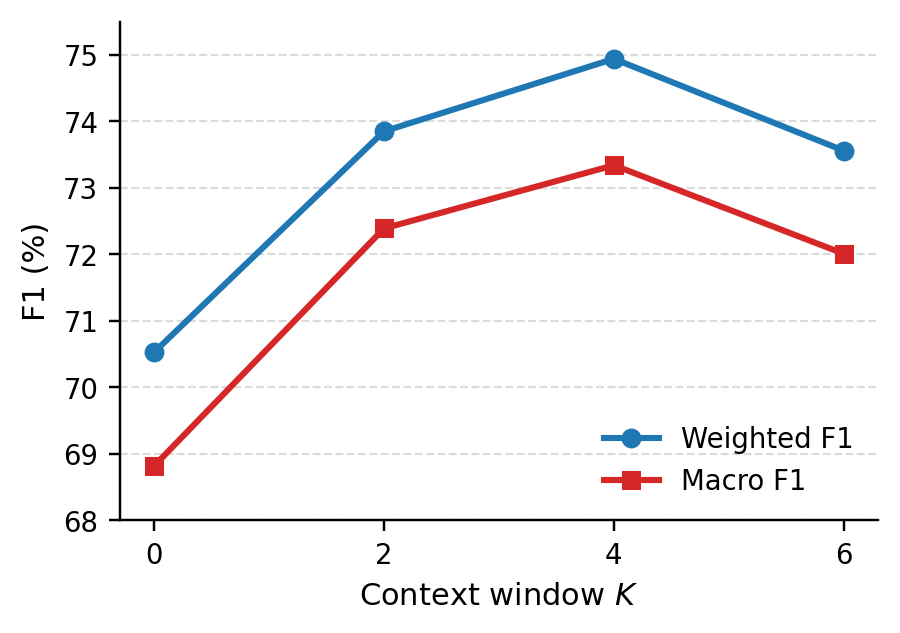}
 \caption{IEMOCAP sensitivity to the number $K$ of preceding dialogue turns.}
 \label{fig:context-window}
\end{figure}

Figure~\ref{fig:context-window} probes conversational-history sensitivity with fixed architecture. Removing history is harmful, but more is not monotonically better: $K=2$ and $K=6$ trail $K=4$. The peak at moderate context suggests diminishing utility as irrelevant turns consume the fixed input budget.

Taken together, the ablations separate expert diversity from routing structure. Uniform averaging tests a constant ensemble; the flat gate retains adaptive weighting but removes hierarchical factorization; and removing the cross-modal expert tests unimodal routing alone. The respective deficits indicate benefits from adaptive weighting and cross-modal interaction, while the 0.34-point hierarchical advantage remains modest.

Class-conditional routing is descriptive rather than causal: it shows variation across emotions, but not that larger weight identifies the objectively cleaner modality. Controlled degradation is needed to test whether operational reliability tracks signal quality.

Together, these analyses qualify the reliability claim. Gate weights are learned proxies for trust, not calibrated measurements of modality quality, and may degrade under unobserved corruption or distribution shift.

\section{Limitations}
\label{sec:limitations}

Evidence is strongest on IEMOCAP; MELD analysis is less complete. Although the matched ablations use three seeds, the small hierarchical-versus-flat difference requires caution. Class-conditional gate averages do not establish causal correspondence to signal quality. Large backbones also confound representation strength with routing, and tests omit controlled missing-modality, acoustic-noise, and transcript-noise conditions.

Graded corruption, missing-modality tests, cross-corpus transfer, and cost-matched backbones would help determine whether routing responds to degraded evidence rather than dataset-specific priors.

\section{Conclusion}
\label{sec:conclusion}

ReH-FUSE treats multimodal ERC as hierarchical routing over contextualized text, audio, and cross-modal decisions. Learned routing outperforms uniform averaging; hierarchical factorization modestly improves on a flat gate in matched IEMOCAP results. These results suggest adaptive selection offers flexibility when modality utility varies. The cross-modal ablation further indicates that routing complements, rather than replaces, explicit interaction between modalities. Future work should test calibration under distribution shift.

\clearpage
\ninept

\section{Compliance with Ethical Standards}
This study used the existing datasets and involved no new data collection from human participants. The datasets were used in accordance with their terms of use.

\section{Acknowledgment}
This work was supported by the NSTC, Taiwan (Grant NSTC 115-2221-E-011-151), and the Empower Vocational Education Research Center at NTUST under the MOE Higher Education Sprout Project. The authors thank the National Center for High-Performance Computing, Taiwan, for computational and storage resources.

\makeatletter
\let\defaultthebibliography\thebibliography
\renewcommand{\thebibliography}[1]{%
  \defaultthebibliography{#1}%
  \setlength{\itemsep}{0pt}%
  \setlength{\parskip}{0pt}%
}
\makeatother
\bibliographystyle{IEEEbib}
\bibliography{reference}

\end{document}